\documentclass[journal]{IEEEtran}
\ifCLASSINFOpdf
  \usepackage[pdftex]{graphicx}
  \usepackage{amssymb}
  \usepackage{amsmath}
  \usepackage[ruled,vlined]{algorithm2e}
  \usepackage{booktabs}
  \usepackage{multirow}
  \usepackage{url}
\else
\fi
\begin{document}
%
\title{SMc2f: Robust Scenario Mining for Robotic Autonomy from Coarse to Fine}
%
%
%


\author{Yifei Chen,~\IEEEmembership{Graduate Student Member,~IEEE,}
        and~Ross~Greer,~\IEEEmembership{Member,~IEEE}
\thanks{Y. Chen is with the Department of Computer Science at Xi'an University of Technology. R. Greer is with the department of Computer Science \& Engineering at the University of California, Merced.}


}

%



\maketitle

\begin{abstract}
The safety validation of autonomous robotic vehicles hinges on systematically testing their planning and control stacks against rare, safety-critical scenarios. Mining these `long-tail' events from massive real-world driving logs is therefore a critical step in the robotic development lifecycle. The goal of the Scenario Mining task is to retrieve useful information to enable targeted re-simulation, regression testing, and failure analysis of the robot's decision-making algorithms. RefAV, introduced by the Argoverse team, is an end-to-end framework that uses large language models (LLMs) to spatially and temporally localize scenarios described in natural language. However, this process performs retrieval on trajectory labels, ignoring the direct connection between natural language and raw RGB images, which runs counter to the intuition of video retrieval; it also depends on the quality of upstream 3D object detection and tracking. Further, inaccuracies in trajectory data lead to inaccuracies in downstream spatial and temporal localization. To address these issues, we propose Robust Scenario Mining for Robotic Autonomy from Coarse to Fine (SMc2f), a coarse-to-fine pipeline that: employs vision–language models (VLMs) for coarse image–text filtering, builds a database of successful mining cases on top of RefAV and automatically retrieves exemplars to few-shot condition the LLM for more robust retrieval, and introduces text–trajectory contrastive learning to pull matched pairs together and push mismatched pairs apart in a shared embedding space, yielding a fine-grained matcher that refines the LLM’s candidate trajectories. Experiments on public datasets demonstrate substantial gains in both retrieval quality and efficiency. We make our code available at: \textit{\url{https://anonymous.4open.science/r/test-EE20}}

\end{abstract}


%
\IEEEpeerreviewmaketitle

\section{Introduction}

The development and validation of autonomous vehicles (AVs) are inherently data-driven endeavors \cite{amini2020learning, forte2024robust}. Modern AV fleets, equipped with a comprehensive suite of sensors—including LiDAR, radar, and high-resolution cameras—generate unprecedented volumes of multi-modal data. A single vehicle can generate terabytes of data per hour during operation. This capacity for large-scale data acquisition presents both an opportunity and a challenge: transforming vast, uncurated driving logs into structured and actionable knowledge is essential for advancing AV safety and reliability. Such raw data is indispensable for training perception models, validating planning and control algorithms, and, most critically, identifying and understanding rare or safety-critical events for targeted system improvement.

A central challenge in this process is the well-documented long-tail problem \cite{nunes2022unsupervised}. The distribution of driving scenarios is highly imbalanced: the majority of collected data corresponds to routine, repetitive events such as highway cruising or waiting at a traffic light. In contrast, complex, unpredictable, and safety-critical scenarios—such as pedestrians unexpectedly crossing against signals, aggressive cut-ins, or interactions with emergency vehicles—are exceedingly rare. Despite their low frequency, these long-tail events are precisely the situations that stress the operational design domain of AVs and are crucial for constructing a comprehensive safety case. However, manually sifting through petabytes of data to extract such rare scenarios is prohibitively time-consuming and costly, making scalable and automated scenario mining a necessity for the AV industry.

Traditional scenario mining approaches \cite{9294652,8906293} typically rely on structured query languages or hand-crafted heuristics to define events of interest. While effective for simple and well-defined events (e.g., identifying left turns at intersections), these methods lack the expressiveness required to capture the nuanced, compositional, and often subjective characteristics of complex real-world interactions. For instance, a query such as “a truck cuts off a car, causing the car to brake abruptly” involves reasoning about causality, temporal ordering, and subjective interpretation—concepts that are difficult to articulate in rigid query-based frameworks.

To address these limitations, the field is rapidly shifting toward natural language–based scenario mining \cite{jung2025gotpr}. This paradigm enables engineers, safety analysts, and developers to describe complex driving situations through intuitive, free-form text. The leading academic benchmark in this area is the Argoverse 2 Scenario Mining Challenge \cite{wilson2021argoverse}, which builds upon the comprehensive Argoverse 2 sensor dataset \cite{wilson2021argoverse}. The challenge includes 1,000 driving logs, each 15 seconds long, collected across six U.S. cities. It features 10,000 diverse, planning-centric natural language queries and requires participating systems to perform a multi-level localization task: (i) determine whether the queried scenario occurs within a log, (ii) temporally localize its start and end times, and (iii) spatially identify and track all relevant actors involved.

The baseline method for this challenge, RefAV \cite{davidson2025refav}, leverages an LLM as a program synthesizer. Specifically, RefAV translates a natural language query into an executable Python program composed of a predefined library of atomic functions. These functions encapsulate fundamental state checks (e.g., is turning()) and relational predicates (e.g., has objects in front()) that operate directly on 3D object tracks. By chaining these functions, the LLM constructs complex logical programs capable of filtering and retrieving trajectory segments that match the semantics of the natural language description.

Despite its contributions, the RefAV framework exhibits several fundamental limitations that hinder both robustness and scalability. One major challenge lies in its reliance on zero-shot code generation by LLMs. Without explicit supervision or task-specific adaptation, LLMs often produce code with syntactic errors, logical inconsistencies, or incorrect parameterizations, which can result in runtime failures that terminate the mining process prematurely. The RefAV inference process operates solely on the abstract outputs of an upstream 3D perception and tracking system, without access to the rich semantic cues embedded in RGB images. Finally, the RefAV pipeline suffers from computational inefficiency. For each query, the LLM-generated program must be executed over the full set of object tracks in a log. This process becomes increasingly time-consuming for complex queries that involve numerous relational checks. These limitations motivate us to develop an alternative approach that establishes a more robust and direct connection between language and visual evidence.

To address these issues, we present Robust Scenario Mining for Robotic Autonomy from Coarse to Fine, a principled multi-stage framework for robust and efficient scenario mining. The design of the framework follows a coarse-to-fine paradigm that progressively narrows the search space while increasing semantic precision.

The main contributions of our research are as follows:

\begin{itemize}

\item [$\bullet$]We employ vision–language models (VLMs) for an initial coarse filtering stage that operates directly on raw RGB video. Leveraging CLIP’s cross-modal alignment, we rapidly identify query-consistent temporal segments, pruning the search space and focusing later stages on the most relevant portions of the log.

\item [$\bullet$]We propose a knowledge-base-guided few-shot method to improve the robustness of code generation. We build a knowledge base of successful description–trajectory–code triplets and retrieve relevant examples for new queries. Integrated into few-shot prompts, these examples provide in-context guidance that enhances code reliability while reducing hallucinations and runtime errors.

\item [$\bullet$]We conduct fine-grained text–trajectory matching with a Dual-Encoder Text–Trajectory Matcher (DETTM). Trained contrastively, DETTM embeds language and spatio-temporal trajectories in a shared space, enabling precise re-ranking of candidate tracks and reducing false positives.

\end{itemize}


\section{Related Work}

\subsection{Video retrieval}
Text–video retrieval is a task closely related to scenario mining: both aim to locate semantically coherent segments within large-scale, long-horizon data streams based on natural language descriptions. Within the video-retrieval literature, multimodal fusion has emerged as the dominant paradigm, offering useful insights for scenario mining.

Since 2020, progress in text–video retrieval has been driven by advances in cross-modal alignment and temporal modeling. In 2020, Gabeur et al. \cite{gabeur2020multi} introduced a multimodal Transformer that jointly encodes visual modalities and models temporal dependencies via cross-modal attention, optimizing language and video features together. In 2021, Wang et al. \cite{wang2021t2vlad} proposed T2VLAD, which employs shared semantic centers to achieve efficient global–local alignment for fine-grained comparisons. In 2022, Gorti et al. \cite{gorti2022x} presented X-Pool, enabling selective attention to semantically relevant frames and effectively filtering visual noise. In 2023, Wu et al. \cite{wu2023cap4video} developed Cap4Video, which leverages zero-shot video-generated captions for data augmentation, cross-modal interaction, and auxiliary inference, achieving strong benchmark results. In 2024, Wang et al. \cite{wang2024text} introduced T-MASS, a stochastic text-embedding strategy that treats queries as deformable semantic masses, incorporating similarity-aware radii and support-text regularization to improve expressiveness, setting new records across five datasets. Entering 2025, Zhang et al. \cite{zhang2025tokenbinder} proposed TokenBinder, a two-stage framework with coarse-to-fine one-to-many alignment and a Focused-view Fusion Network for cross-attention, attaining state-of-the-art results across six benchmarks, while Bian et al. \cite{bian2025selective} released the SMA framework, which performs selective multi-grained alignment at both video–sentence and object–phrase levels using token aggregation and similarity-aware keyframe selection, yielding strong performance on MSR-VTT, ActivityNet, and beyond.

The motivation behind these advances closely parallels that of scenario mining: both tasks require locating a contiguous scene that corresponds to a natural-language description. However, scenario mining often demands finer-grained retrieval focused on the trajectory of a specific agent or a set of interacting agents. Whereas most video-retrieval approaches rely on direct alignment between visual and textual semantics, this presents an opportunity for scenario mining: cross-modal semantic matching can serve as a coarse filtering stage, after which detailed analysis can be restricted to the subset of trajectories identified.
\begin{figure}[t]
    \centering
    \includegraphics[width=1\linewidth]{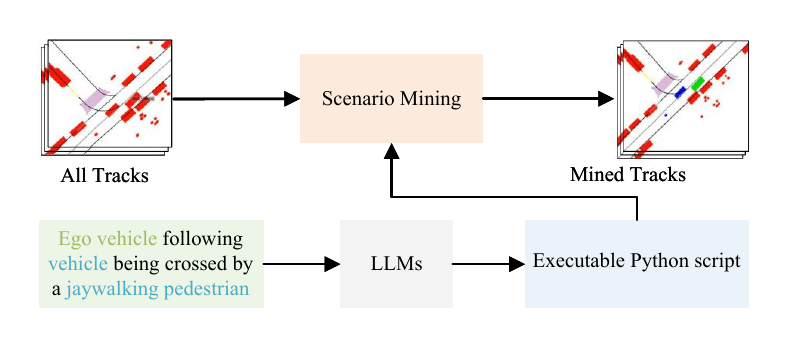}
    \caption{Overview of the RefAV baseline framework. A natural language query is processed by a Large Language Model (LLM) to synthesize an executable Python script. This script then filters the full set of trajectory data (``All Tracks") to retrieve specific segments that match the scenario description (``Mined Tracks").}
    \label{fig2}
\end{figure}
\subsection{Scenario mining}

Early methods in scenario mining focused on rule-based systems and human-defined criteria. These methods encode traffic laws, domain expertise, and parameters from accident databases into formal languages and ontologies \cite{li2021vehicle}. By defining a logical scenario space with parameters (e.g., road curvature, number of vehicles, weather) and their valid ranges, scenarios can be generated through techniques like combinatorial testing to cover a wide array of predefined conditions. A significant initiative in this area is the PEGASUS project \cite{PegasusScenario2017}, which established a systematic, knowledge-driven workflow for defining scenarios. Formal languages such as ASAM OpenSCENARIO \cite{ASAMOpenSCENARIO2022} have become industry standards for describing the dynamic content of driving scenarios. The primary advantage of knowledge-based methods is the high degree of control and interpretability, making them ideal for testing system compliance with known rules. However, their main limitation is that they are bound by existing knowledge and manual effort, often failing to uncover novel modes and lacking the behavioral complexity of real-world traffic.

With the advent of large-scale, real-world driving datasets, data-driven methods have become prominent. These approaches mine sensor logs to extract realistic scenarios or learn generative models of traffic behavior. A typical pipeline acquires data from the Waymo Open Motion Dataset \cite{sun2020scalabilityperceptionautonomousdriving}, nuScenes \cite{caesar2020nuscenesmultimodaldatasetautonomous}, or Argoverse 2 \cite{wilson2023argoverse2generationdatasets}, identifies high-criticality events, and—via SQL, a bespoke DSL, or a general-purpose language—recasts scenario search as label retrieval. Erwin de Gelder et al. \cite{Gelder_2020} demonstrate a label-based system on pre-annotated datasets; however, such labels are coarse and scale poorly. Motional’s pipeline \cite{motional_scenario_mining_2022} follows a continual-learning loop, maintaining a tag vocabulary that encodes basic spatio-temporal relations and storing labels in a relational database for efficient retrieval. While data-driven methods yield realistic, behavior-grounded scenarios, limited coverage makes rare edge cases a “needle in a haystack,” and outputs are often descriptive rather than adversarially challenging. To overcome these passive-mining limits, recent work applies large foundation models \cite{wang2025generativeaiautonomousdriving, liang2024aide}. As surveyed by Gao et al. \cite{gao2025foundationmodelsautonomousdriving}, LLMs, VLMs, and diffusion models enable scenario generation from high-level prompts—for example, “Create a challenging scenario where a truck illegally overtakes a bicycle on a rainy night”—and produce simulator-ready parameters.

The work by Manmohan Chandraker et al. \cite{khan2022single} initially focused on vision--language alignment tasks, and subsequently extended similar ideas to data mining in autonomous driving. They proposed an automated and iterative data engine for autonomous driving \cite{liang2024aide} that realizes a complete closed-loop pipeline, including issue identification, data mining, automatic labeling, verification, and iterative model improvement.
Works such as ChatScene \cite{zhang2024chatsceneknowledgeenabledsafetycriticalscenario} have demonstrated the ability of LLMs to understand complex spatial and behavioral relationships to produce diverse and contextually rich scenarios. This approach holds immense promise for bridging the gap between abstract human knowledge and concrete simulation data. While still an emerging area, the key challenges include ensuring the physical plausibility and controllability of generated scenarios and managing the significant computational resources required by these large models. RefAV \cite{davidson2025refav} is a large-scale scenario-mining framework that houses 10,000 distinct natural-language queries describing the complex multi-agent interactions present in the 1,000 driving logs of the Argoverse 2 sensor suite. It exposes 28 handcrafted atomic functions capable of recognizing trajectory states, expressing relational predicates between a target agent and its surrounding entities, and supporting basic Boolean logic. At its core, the framework feeds the natural-language query, the atomic function inventory, and carefully engineered prompts into an LLM, which synthesizes an executable script composed of these atoms; running the script over the dataset then retrieves the trajectories that satisfy the query, as shown in Fig. \ref{fig2}. Zheng et al. \cite{chen2025smagent} optimize atomic function generation via independent LLM agents, while Chen et al. \cite{chen2025technicalreportargoverse2scenario} further boost RefAV through a fault-tolerant design coupled with more effective prompt engineering. Building on RefAV, we decouple Scenario Mining into a coarse-to-fine pipeline by combining multimodal semantic filtering, few-shot LLM prompting, and a fine-grained text–trajectory matcher, achieving gains in both efficiency and accuracy.


\begin{figure*}[!t]
    \centering
    \includegraphics[width=1\textwidth,keepaspectratio]{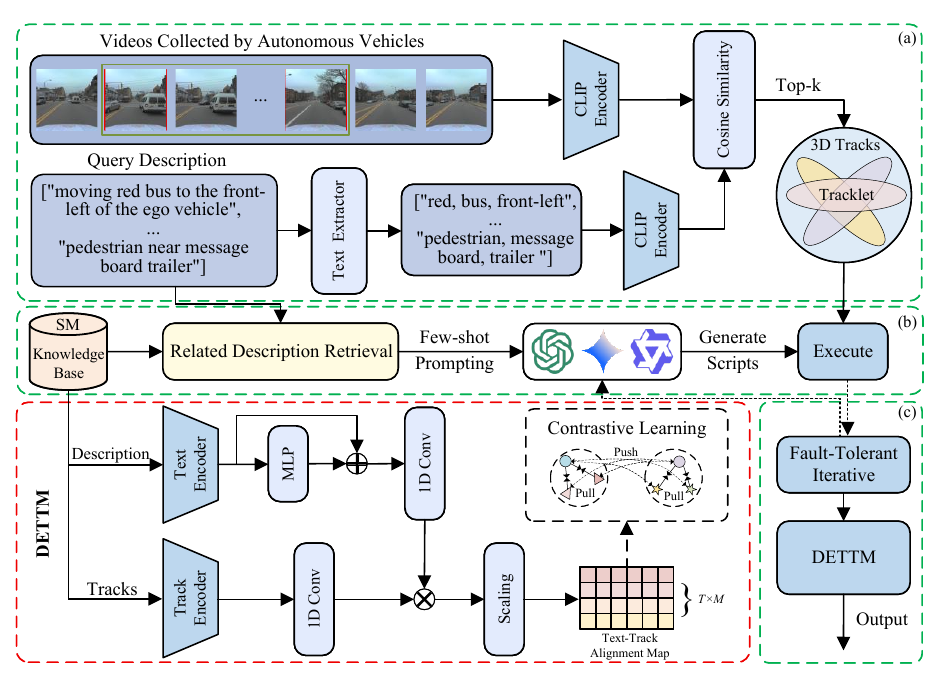}
    \caption{The framework of SMc2f. The boxes in video frames represent randomly sampled frames within the sliding window; the red dashed lines indicate training, and the green dashed lines indicate inference.}
    \label{fig-big}
\end{figure*}

\section{Method}
Fig. \ref{fig-big} overviews our framework, comprising three components: (a) a CLIP-based multimodal coarse filter, (b) knowledge-base-guided code generation for scenario mining, and (c) a contrastive-learning-driven trajectory fine filter. Specifically, to make the entire scenario-mining process fast and accurate, and to compensate for the lack of RGB utilization in the RefAV pipeline, we extract image features for all video frames and text features for the query description, select segments with higher cosine similarity, and restrict precise mining to the subset of trajectories whose timestamps align with these segments. We build a scenario-mining knowledge base from prior successful queries, storing triples of \{natural-language description, associated trajectories, executable Python script\}. At inference time, few-shot examples from the knowledge base condition the LLM, improving its handling of atomic functions involving spatial relationships. We further train a fine-grained text–trajectory matcher via contrastive learning on knowledge-base pairs; the candidate trajectories surfaced by component (b) are scored by this matcher to perform fine-level filtering.

\subsection{CLIP-based Coarse Filtering}
Within the original RefAV pipeline, scenario mining is performed exclusively via scripts assembled from atomic functions, a design that overlooks the direct correspondence between natural-language queries and raw image frames. In RefAV, the pipeline begins with 3D object detection to extract each target’s class, heading, velocity, and related attributes, assigns corresponding labels, and then waits for an LLM-generated script to query them. The correctness of those attributes rests wholly on the upstream detector’s performance, and the elongated processing chain renders the connection between natural-language queries and raw sensor data highly indirect—raising the likelihood that crucial information will be missed or inaccurately captured.

Inspired by advances in vision–language modeling, we redesign the pipeline with a VLM-based coarse filtering stage that directly compares natural-language queries with raw RGB imagery, thereby temporally localizing the candidate regions before any symbolic reasoning is invoked. Concretely, each Argoverse~2 log spans 15s of synchronized sensor data captured by seven ring cameras at 20 Hz. We apply a sliding window of length $W$ seconds and stride $S$ seconds to partition the log into overlapping temporal clips. From each window we sparsely sample $n$ frames, forming a compact visual summary that is later mean-pooled. To enable efficient query-time retrieval, we precompute per-frame embeddings $f_{\text{img}}\in\mathbb{R}^{d}$ across the entire corpus using a pretrained CLIP image encoder (e.g., ViT-L/14). These are stored offline, so that no expensive vision inference is performed at query time.

Given a natural-language query, we employ an industrial-grade NLP library, spaCy, to extract the colour, entity, and spatial-relation words from each query, then encode it with the CLIP text encoder to obtain $f_{\text{text}}\in\mathbb{R}^{d}$. The sampled frame embeddings of each window are aggregated into a window-level representation:
\begin{equation}
f_{\text{win}} = \frac{1}{n} \sum_{i=1}^{n} f_{\text{img}, i}.
\end{equation}
Window relevance is then measured by cosine similarity:
\begin{equation}
\mathrm{sim}\left(f_{\text{text}}, f_{\text{win}}\right) =
\frac{f_{\text{text}} \cdot f_{\text{win}}}
{\lVert f_{\text{text}}\rVert ,\lVert f_{\text{win}}\rVert}.
\end{equation}
We rank the windows by the accumulated similarity scores across all seven views, retain the top-$k$ results, and merge any temporally adjacent windows to form a reduced search region. Only trajectories that overlap this region are forwarded to the subsequent LLM-based mining stage. This hierarchical approach prunes irrelevant frames early, significantly shrinking the search space and lowering computation while preserving recall. Executing the original RefAV scripts within this semantically aligned subset markedly improves the HOTA-T metric and reduces inference cost.

\subsection{Scenario Mining Knowledge Base}
A central component of SMc2f is the Scenario Mining Knowledge Base, a structured base designed to mitigate the brittleness of zero-shot LLM code generation. Rather than relying solely on the model’s improvisation, the knowledge base functions as a form of accumulated experience, supplying the LLM with relevant and validated examples to guide program synthesis. This design follows the Retrieval-Augmented Generation (RAG) paradigm \cite{ahmed2022fewshottrainingllmsprojectspecific}, which has proven effective in reducing hallucinations and improving factual grounding in large language models.


The knowledge base is built in a one-time offline process using the official training split of the Argoverse 2 Scenario Mining benchmark. Each natural language query and its associated ground-truth tracks are paired with a candidate Python script generated from the RefAV atomic function library. Script generation is semi-automated: a powerful LLM (Gemini 2.5 pro) produces initial code, which is then executed on the log and strictly validated against the ground truth. Only scripts with perfect matches are accepted. When verification fails, the script is refined—either through error-guided re-prompting or manual correction—and re-tested until fidelity is ensured. The resulting triplets of query, trajectory, and code are indexed, with queries pre-encoded into dense vectors to enable fast semantic retrieval. Within the SMc2f pipeline, the knowledge base plays a dual role. During Stage 2, it serves as the retrieval pool for automated prompting, allowing relevant examples to be dynamically incorporated into few-shot prompts that improve LLM reliability. During Stage 3, its verified query–trajectory pairs provide the positive examples needed to train the contrastive text–trajectory matcher. In this way, the same resource that stabilizes code generation also underpins fine-grained retrieval, aligning the two stages into a coherent and robust framework.

\subsection{Few-Shot Prompting for Robust Code Generation}
RefAV feeds atomic functions and prompts into an LLM and asks it to generate executable code to mine scenarios that match the description. In practical inference, this process often suffers from unstable mining quality and incorrect use of atomic functions. For example, under zero-shot settings the LLM frequently misuses relative spatial parameters in the atomic functions, often swapping the subject and object of spatially related arguments. Fundamentally, the LLM lacks exemplar-based guidance. We find that when complete cases containing both code and prompts are provided as guidance, the robustness of the LLM’s code generation improves markedly, along with its ability to reason about relative spatial relations.

When a new query, $q_{\text{new}}$, is presented, we first retrieve the most semantically relevant examples from our knowledge base $K$. This is accomplished using a pre-trained sentence encoder, Sentence-BERT \cite{reimers2019sentencebertsentenceembeddingsusing}, which is optimized for mapping sentences to a dense vector space where semantic similarity is captured by cosine distance. We precompute and index the sentence embeddings for all queries $q_i$ in $K$. At inference time, we encode the new query $q_{\text{new}}$ to obtain its embedding $e_{\text{new}}$. We then perform an efficient KNN \cite{Cunningham_2021} search against the indexed embeddings to retrieve the top-$k$ most similar queries and their corresponding code snippets, yielding a set of exemplar pairs: $\{(q_j, C_j)\}_{j=1}^{k}$. These retrieved exemplars are dynamically formatted into a few-shot prompt for LLM.

\begin{figure}[!h]
    \centering
    \includegraphics[width=1\linewidth]{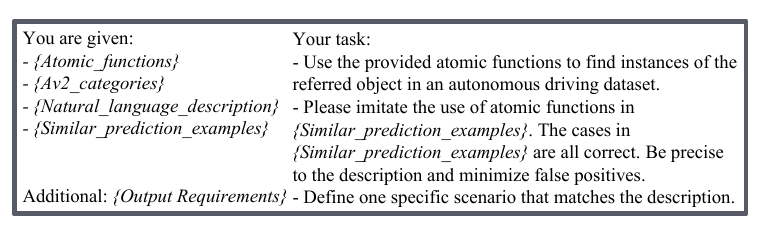}
    \caption{Few-shot prompting instructions provided to LLMs.}
    \label{prompt}
\end{figure}

To provide the LLM with complete context for better understanding, we perform prompt engineering augmented by retrieving relevant descriptions from the knowledge base. The prompt design consists of three parts: (i) Inputs: atomic functions, Argoverse 2 object categories, the natural-language query, and topically similar exemplars retrieved from the knowledge base; (ii) Instruction: an explicit role and objective for the LLM, together with a request to pattern its solution after the retrieved exemplars; and (iii) Output requirements: code style guidelines and a constrained response format. The full prompt content is shown in Fig. \ref{prompt}.

LLM-generated code often fails at runtime, prematurely aborting scenario mining and reducing coverage and reliability. Upon receiving the prompt, the LLM generates Python code that composes the available atomic functions. We execute the candidate; if it completes without error, the pipeline terminates. Otherwise, on any exception (e.g., \texttt{NameError}, \texttt{TypeError}), we record the error string and return both the failing code and message to the LLM with an explicit instruction to correct the fault. The model produces a revised candidate, which we re-execute. This loop repeats until the program runs or the iteration budget $K'$ is exhausted. In our setup, $K'{=}5$; queries still unresolved after five attempts are flagged for manual review, preventing unbounded retries and improving overall reliability.







\subsection{Fine Filtering Based on Contrastive Learning Matcher}
The fine filter aims to refine the candidate tracks produced in step (b) by selecting those that more precisely satisfy the language description. We consider paired data $\mathcal{D}=\{(x_i, y_i)\}_{i=1}^{N}$, where $x_i$ is an object track and $y_i$ is a natural-language description. Our objective is to learn cross-modal representations that enable global retrieval and fine-grained track–description alignment.

All training data are from the Scenario Mining Knowledge Base. We extract the description–track components from each triple and use them as training data. The CLIP text encoder encodes the descriptions as text features, and a pre-trained PatchTST \cite{nie2023timeseriesworth64}—due to its temporal modeling ability and built-in temporal positional embeddings—is used to extract features from the tracks. The two feature types are aligned via an alignment map and optimized with contrastive learning. Description–track pairs that match in the knowledge base are treated as positive samples, while a description paired with other tracks constitutes negative samples. Training pulls positive pairs closer and pushes negative pairs apart in the embedding space, thereby achieving fine-grained filtering of tracks.

Specifically, we treat each track as a length-$L_i$ sequence of raw, native Argoverse 2 fields; at time $t$:
\begin{equation}
    u_{i,t} = [\, t_x,\, t_y,\, t_z,\, q_w,\, q_x,\, q_y,\, q_z,\, \ell,\, w,\, h \,] \in \mathbb{R}^C, \quad C=10,
\end{equation}
where $t_x,t_y,t_z$ denotes the object's 3D translation in the ego-vehicle coordinate system and $q_w,q_x,q_y,q_z$ are the unit quaternion encoding the orientation of the object and $\ell,w,h$ represent the size of the 3D bounding box oriented: length, width and height. Stack $\hat{u}_{i,t}$ into $\hat{U}^i \in \mathbb{R}^{L_i \times C}$ which denote the $C$-dimensional, per–time step feature vector for sample $i$ at time index $t \in \{0, \ldots, L_i - 1\}$. To ensure numerical stability and account for the disparate scales across different physical quantities, we apply z-score normalization to each feature dimension across the training set before feeding them into the encoder. PatchTST slices the time axis with patch length $P$ and stride $S$, producing

\begin{equation}
    B_i^{(0)} \in \mathbb{R}^{T_i \times d_0}, 
\quad T_i = \left\lfloor \frac{L_i - P}{S} \right\rfloor + 1,
\end{equation}
where each token summarizes one time patch. Patch-level positional encodings are applied internally. $B_i^{(0)}$ is the token sequence produced by applying a learnable linear map to each $P \times C$ time slice, implemented as a 1D convolution along the time axis with kernel size $P$, stride $S$, input channels $C$, and output channels $d_0$ followed by adding patch-level positional encodings. $T_i$ is the resulting number of time patches kept when only complete patches are used. Meanwhile, the descriptions are encoded with the CLIP text encoder. Their features are then processed by a two-layer MLP with skip connections and passed through a 1D convolution, which provides stronger local pattern modeling and adaptability for sequential inputs. Then, the fine cross-modal correspondences between track patches and text tokens are captured by:
\begin{equation}
    S = \frac{(B W_q)(A W_k)^\top}{\sqrt{d_k}} \in \mathbb{R}^{T \times M},
\end{equation}
where $B$ represents the track token matrix, $A$ represents the text token matrix and the $W_q/W_k$ represent the learnable projections to a shared query/key space of size $d_k$. We employ a Multiple Instance Learning (MIL) loss to constrain and optimize the alignment between track patches and text tokens. This objective encourages the model to assign high similarity to the correct local correspondences, thereby improving sequence-level discriminability, as shown in Eq. \ref{loss1}:
\begin{equation}
\mathcal{L}_{\mathrm{MIL}} 
= \frac{1}{N} \sum_{i=1}^{N} 
\left[
- \log 
\frac{\exp\!\left( z_i / \gamma \right)}
{\exp\!\left( z_i / \gamma \right) + \sum_{j \neq i} \exp\!\left( z_{ij}^- / \gamma \right)}
\right],
\label{loss1}
\end{equation}
where we condense the alignment matrix into a single evidence score $z$ that reflects how strongly a track aligns with a text at the most relevant moments. The overall contrastive objective is guided by a global symmetric InfoNCE loss, which pulls each matched track–text pair together in the embedding space while pushing mismatched pairs apart. By enforcing both retrieval directions, it improves bidirectional retrieval performance and stabilizes training. Given L2-normalized global embeddings $\hat{b}^i, \hat{a}^i \in \mathbb{R}^e$ for track $i$ and text $i$ in a minibatch of size $N$, define the scaled similarities as:

\begin{equation}
    s_{ij} = \frac{(\hat{b}^i)^\top \hat{a}^j}{\tau}, \quad \tau > 0.
\end{equation}

The global symmetric InfoNCE loss function jointly optimizes text→track and track→text retrieval, with all other items in the batch serving as in-batch negatives, as shown in Eq. \ref{loss2}:

\begin{IEEEeqnarray}{rCl}
\mathcal{L}_{\text{Global}} 
&=& \tfrac{1}{2} \Biggl[
\frac{1}{N} \sum_{i=1}^{N} 
- \log \frac{\exp(s_{ii})}{\sum_{j=1}^{N} \exp(s_{ij})} \nonumber \\
&&\;+\; \frac{1}{N} \sum_{i=1}^{N} 
- \log \frac{\exp(s_{ii})}{\sum_{j=1}^{N} \exp(s_{ji})}
\Biggr].
\label{loss2}
\end{IEEEeqnarray}

Overall, the optimization object can be formulated as:
\begin{equation}
    \mathcal{L} = \lambda_1 \mathcal{L}_{\text{MIL}} + \lambda_2 \mathcal{L}_{\text{Global}},
\end{equation}
where $\lambda_1,\lambda_2$ denotes the hyperparameter.

\section{EXPERIMENTS}

\subsection{Dataset and Evaluation Metrics}
The experiments were conducted using the Argoverse 2 dataset. The dataset provides rich multi-modal information, including RGB camera frames, LiDAR point clouds, HD Maps, and 3D track annotations for 26 object categories. Argoverse 2 is currently the only dataset available for scenario mining task, and RefAV remains the sole benchmark built upon it.

The primary metric is HOTA-Temporal \cite{davidson2025refav}. It is a spatial tracking metric that considers only the scenario-relevant objects during the precise timeframe when the scenario is occurring. 
HOTA \cite{Luiten_2020} was introduced to provide a unified evaluation of multi-object tracking by jointly accounting for detection, association, and localization—three facets that together reflect human intuition of tracking quality.
Secondary metrics include HOTA, Timestamp F1, and Log F1. Timestamp F1 treats the video as a sequence of frames, labeling each timestamp as “scenario” or “non-scenario.” Precision and recall are computed from the comparison of predicted and ground-truth frame labels. Log F1 simplifies the task to a single binary decision per log. After aggregating true positives, false positives, and false negatives across all logs, a conventional F1-score is produced. 

\subsection{Implementation Details}
Our entire framework is implemented using PyTorch framework, with all experiments conducted on a workstation equipped with two NVIDIA GeForce RTX 4090 GPUs. For the initial coarse filtering stage, we employ the pre-trained ViT-L/14 CLIP model. Each 15-second log is partitioned into temporal clips using a sliding window of length $W=3$ seconds and a stride of $S=1$ second, from which we uniformly sample $n=5$ frames per camera view. The top $k=5$ most relevant windows, ranked by cosine similarity, are selected to narrow the search space. In the second stage, robust code generation is driven by the Gemini 2.5 Pro large language model, configured with a low temperature of $0.2$ to ensure deterministic output. To construct the few-shot prompts, we retrieve the top $10$ semantically similar examples from our knowledge base using a KNN search on embeddings generated by a Sentence-BERT model; the subsequent fault-tolerant execution loop attempts to correct runtime errors for a maximum of $K'=5$ iterations. Finally, for fine-grained filtering, our Dual-Encoder Text–Trajectory Matcher (DETTM) is trained from the verified pairs in the knowledge base. The DETTM's track encoder is a PatchTST model with 3 transformer layers, 8 attention heads, and a 256-dimensional embedding space, processing track sequences with a patch length of $P=16$ and a stride of $S=8$. The text encoder utilizes the frozen text tower from the same `ViT-L/14' CLIP model. Both modalities are projected into a 512-dimensional shared embedding space. We train the matcher for 50 epochs with a batch size of 128, using the AdamW optimizer with a learning rate of $1 \times 10^{-4}$ and weight decay of $0.01$, managed by a cosine annealing scheduler with a 5 epoch warm up. The MIL and global InfoNCE loss weights are set equally ($\lambda_1=1.0$, $\lambda_2=1.0$), with temperature parameters of $\gamma=0.1$ and $\tau=0.07$, respectively.

\subsection{Results}
To evaluate the overall performance of our proposed SMc2f framework, we compare it against the RefAV baseline on the Argoverse 2 validation set. We use the superior Le3DE2E and TransFusion method as the 3D tracking backbone for both frameworks to ensure a fair comparison of the scenario mining logic itself. As shown in Table \ref{tb1}, our full coarse-to-fine pipeline achieves substantial improvements across all evaluation metrics. The $^{*}$ represents the baseline reproduced in our implementation.
\begin{table}[t]
\centering
\setlength{\tabcolsep}{3pt}
\caption{Comparison with SOTA Methods on the Argoverse2 Dataset}
\begin{tabular}{cccccc}
\toprule
\textbf{3D Track} & \textbf{Model} & \textbf{HOTA-T} & \textbf{HOTA} & \textbf{TS-F1} & \textbf{Log-F1} \\
\midrule
\multirow{3}{*}{Le3DE2E} 
  & RefAV$^{*}$ (Baseline) \cite{davidson2025refav} & 40.17 & 40.33 & 66.70 & 62.71 \\
  & Chen et al. \cite{chen2025technicalreportargoverse2scenario}        & 46.71 & 45.93 & 72.30 &  61.36 \\
  & SMc2f (Ours) & \textbf{53.12} & \textbf{54.05} & \textbf{76.55} & \textbf{75.89} \\
\midrule
\multirow{3}{*}{TransFusion} 
  & RefAV$^{*}$ (Baseline) \cite{davidson2025refav} & 35.50 & 35.93 & 59.89 & 59.13 \\
  & Chen et al.$^{*}$ \cite{chen2025technicalreportargoverse2scenario}        & 40.11 & 41.19 & 62.37 & 61.70 \\
  & SMc2f (Ours) & \textbf{51.89} & \textbf{51.72} & \textbf{74.04} & \textbf{73.41} \\
\bottomrule
\end{tabular}
\label{tb1}
\end{table}
Our method, SMc2f, demonstrates a remarkable performance leap, achieving a score of 53.12 on HOTA-T, which is a 12.95-point absolute improvement over the baseline. Similar significant gains are observed in HOTA (+13.72), TS-F1 (+9.85), and Log-F1 (+13.18). This highlights the effectiveness of our multi-stage design, where initial VLM-based filtering, robust LLM-guided code generation, and fine-grained trajectory matching work in synergy. The substantial increase in Log-F1 confirms our ability to accurately identify relevant logs, while the large gains in HOTA-T and HOTA underscore the precision of the spatio-temporal localization of the final retrieved tracks.

For scenario mining systems to be viable in production, where they must process petabytes of data, inference efficiency is as crucial as accuracy. We perform a rigorous efficiency analysis to quantify the performance gains of our architectural choices. We benchmark two methods, the RefAV* baseline and our full SMc2f framework, across several state-of-the-art large language models (LLMs) to demonstrate the efficiency improvements. The total wall-clock time required to process a single query-log pair was measured on a NVIDIA RTX 4090 GPU. The results highlight a substantial and consistent reduction in inference time, as shown in Table \ref{tab_e}.

\begin{table}[t]
\centering
\caption{Efficiency Comparison Across LLM Backends}
\renewcommand{\arraystretch}{0.9} 
\setlength{\tabcolsep}{3pt}       
\begin{tabular}{llcc}
\toprule
\textbf{LLMs} & \textbf{Method} & \textbf{Time (s) ↓} & \textbf{Relative Speedup ↑} \\
\midrule
\multirow{2}{*}{Qwen2.5-VL-7B} 
 & RefAV* (Baseline) & 41.6 & 1.0$\times$ \\

 & SMc2f (Ours) & \textbf{21.1} & \textbf{1.97$\times$} \\
\midrule
\multirow{2}{*}{GPT 4} 
 & RefAV* (Baseline) & 44.9 & 1.0$\times$ \\
 & SMc2f (Ours) & \textbf{22.7} & \textbf{1.98$\times$} \\
 
\midrule
\multirow{2}{*}{Gemini 2.5 Pro} 
 & RefAV* (Baseline) & 42.7 & 1.0$\times$ \\

 & SMc2f (Ours) & \textbf{24.4} & \textbf{1.75$\times$} \\

\bottomrule
\end{tabular}
\label{tab_e}
\end{table}

\subsection{Ablation Study}

We conduct a thorough ablation study to dissect the contribution of each key component within our SMc2f framework. We evaluate different combinations of our three core modules on top of the RefAV baseline: the CLIP-based Coarse Filter, the Knowledge Base and Few-Shot prompting module (KB+FS), and the final contrastive Fine Filter (FF). The results, presented in Table \ref{tab2}, validate our design choices and demonstrate the synergistic effect of our pipeline.

\begin{table}[t]
\centering
\caption{Ablation Study}
\renewcommand{\arraystretch}{0.9} 
\setlength{\tabcolsep}{3pt}      
\begin{tabular}{ccccc}
\toprule
\textbf{Method} & \textbf{HOTA-T} & \textbf{HOTA} & \textbf{TS-F1} & \textbf{Log-F1} \\
\midrule
Baseline (RefAV*)              & 40.17 & 40.33 & 66.70 & 62.71 \\
+ Coarse Filtering (CF)        & 48.30 & 49.52 & 72.30 & 73.41 \\
+ CF + KB \& Few-Shot (KB+FS)  & 51.25 & 52.18 & 75.02 & 74.95 \\
+ CF + KB+FS + Fine Filter (Ours)& \textbf{53.12} & \textbf{54.05} & \textbf{76.55} & \textbf{75.89} \\
\bottomrule
\end{tabular}
\label{tab2}
\end{table}

Introducing the CLIP-based filter provides the single largest performance boost. It improves HOTA-T by 8.13 points and Log-F1 by a massive 10.7 points. This confirms our hypothesis that directly leveraging visual cues to prune the search space is highly effective, eliminating irrelevant temporal segments and allowing the subsequent stages to focus on promising candidates. By augmenting the LLM with a knowledge base and few-shot examples, we achieve another significant gain, particularly in HOTA-T (+2.95) and TS-F1 (+2.72). This demonstrates that guiding the LLM with successful precedents improves the quality and robustness of the generated code, leading to more precise retrieval and fewer errors from misinterpreting complex spatial relationships. The final addition of our contrastive learning-based matcher provides a further refinement, boosting all metrics, with HOTA-T increasing by another 1.87 points. This component acts as a final verification step, re-ranking the candidate trajectories generated by the LLM script and filtering out subtle false positives.

\section{Conclusion}


We present SMc2f, a coarse-to-fine framework that mines safety-critical scenarios from fleet-scale robot logs. A vision–language filter localizes time windows, a knowledge-base-guided LLM generates auditable programs, and a contrastive text–trajectory matcher performs fine ranking. On Argoverse 2, SMc2f surpasses RefAV on all metrics. Ablations confirm each component’s contribution and robustness to perception noise, enabling scalable, simulator-ready tests for autonomous robots.



%






\IEEEtriggercmd{\enlargethispage{-5in}}


\bibliographystyle{IEEEtran}
\bibliography{IEEEabrv}
\end{document}